\documentclass[11pt,a4paper]{article}
\usepackage[hyperref]{acl2020}
\usepackage{times}
\usepackage{footnote}
\usepackage{latexsym}
\usepackage{xcolor}
\usepackage{url}
\usepackage{graphicx} 
\usepackage{booktabs}
\usepackage{multirow}
\usepackage{amsmath}
\usepackage{color, soul}

\usepackage{pbox}
\usepackage{enumitem}
\usepackage{arydshln}
\setlist[itemize]{leftmargin=*}\setlist{nosep}

\usepackage{microtype}

\aclfinalcopy 

\title{\textsc{orgFAQ}: A New Dataset and Analysis on Organizational FAQs and User Questions}

\author{Guy Lev, Michal Shmueli-Scheuer, Achiya Jerbi\thanks{This work was done while working at IBM.} , David Konopnicki\\
  IBM Research AI - Haifa \\
  {\{guylev,shmueli,davidko\}@il.ibm.com}, {achiyaj@gmail.com}}

\date{}

\begin{document}
\maketitle
\begin{abstract}
Frequently Asked Questions (FAQ) webpages are created by organizations for their users. FAQs are used in several scenarios, e.g., to answer user questions. On the other hand, the content of FAQs is affected by user questions by definition.
In order to promote research in this field, several FAQ datasets exist. 
However, we claim that being collected from community websites, they do not correctly represent challenges associated with FAQs in an organizational context. Thus, we release \textsc{orgFAQ}, a new dataset composed of $6988$ user questions and $1579$ corresponding FAQs that were extracted from organizations' FAQ webpages in the Jobs domain. In this paper, we provide an analysis of the properties of such FAQs, and demonstrate the usefulness of our new dataset by utilizing it in a relevant task from the Jobs domain. We also show the value of the \textsc{orgFAQ} dataset in a task of a different domain - the COVID-19 pandemic.
\end{abstract}

\section{Introduction}
\label{intro}
FAQs are used by organizations (e.g., enterprises, governments, educational institutions) in several scenarios, e.g., to answer user questions in websites or bots interactions. On the other hand, the content of FAQs is affected by user questions by definition.
In customer support scenarios, FAQs help reduce support agents workload, improve user experience and more. In fact, FAQ pages are used by $73\%$ of users\footnote{\url{https://tinyurl.com/uk9qwv2}}. 

FAQs are usually formalized by organizations' experts and reflect the organizational voice. They are well-written 
and focused. We refer to those FAQs as {\it organizational FAQs}, i.e. organizational questions and answers.

Questions written by users w.r.t. services and products appear on the web in various contexts: questions are being sent to search-engines and bots, posted on web forums and sent by email. We refer to those questions as {\it user questions}. 
Specifically, in web forums, such as Yahoo! Answers and Reddit, user questions are answered by other users or, sometimes, by organizations' representatives. Selected questions and answers from these forums constitute {\it community FAQs} (community questions and answers), as opposed to the official organizational FAQs. Note that the community questions are a subset of user questions. 

\begin{figure}[t]
  {\footnotesize
	\setlength{\fboxsep}{4pt}
	\fbox{
		\parbox{0.95\columnwidth}{
				
     \textbf{User question:} \textit{Oh no! I’ve spilled wine over my favourite rug and the stain is not coming off! What should I do?}
     
     \textbf{Community question:} \textit{how do i remove a red wine spillage on our new carpet?!! thanx?}
     
     \textbf{Organizational question:} \textit{How to remove a wine stain from carpet?}
    	}}
  }
  \caption{Examples of user question, community FAQ, and an organizational FAQ, on ``wine stain removal'' issue.}
	\label{fig:example}
\end{figure}

Several research FAQ datasets exist, based on data extracted from web forums. Older datasets like \textsc{Usenet} FAQ\footnote{\url{https://tinyurl.com/wsywgpg}}, and Game FAQ\footnote{\url{www.gamefaqs.com/}}
only contain FAQs (questions and answers). Recent datasets as \textsc{FAQIR}~\cite{FAQIR} and \textsc{StackFAQ}~\cite{stackFAQ} associate FAQ questions with additional user questions having similar meaning. Those user questions are either extracted from forums or created using crowdsourcing. Those richer datasets can be used 
for different tasks like question answering, text generation and more.

Figure~\ref{fig:example} shows a user question and a related community FAQ\footnote{For clarity, in this paper, the term {\it{FAQ}} refers only to the question, while the answer is explicitly referred to as the \it{FAQ answer}.}, both from
the \textsc{FAQIR} dataset, along with a relevant organizational FAQ from a cleaning-products company.
The community FAQ contains uncommon terminology (``wine spillage'' vs. ``wine stain''),
extra information (``new carpet'' vs. ``carpet''), grammatical errors (``on'' vs. ``from''), etc.

We argue that the language used in community FAQs and user questions is different from the language of organizational FAQs. Hence, existing FAQ datasets, which are based on community FAQs, are unsuitable for some tasks: for example, question answering using organizational FAQs, or updating organizations' FAQ pages so they are in tune with users' current questions. In those cases, the language mismatch between user questions and organizational FAQs is an issue not addressed by current datasets.

Hence, our main contributions are as follows: (1) we study the difference between organizational FAQs, community FAQs and user questions; (2) we create and publish a new dataset, \textsc{orgFAQ} (acronym for organizational FAQ)
, containing $6988$ user questions
mapped to $1579$ organizational FAQs from the
Jobs domain\footnote{Data will be available upon paper acceptance.}; (3) we demonstrate the usefulness of this dataset by automatically generating organizational FAQs given user questions, and evaluating the results for the Jobs domain as well as for the urging COVID-19 pandemic domain.
To our knowledge, this is the first work that studies these important aspects of organizational FAQs.

\section{Related Work}
\label{related}

Many works on FAQs focus on the question answering task,
usually given user queries. Early works~\cite{faqfinder,faqfinder1} use semantic features to match questions and answers.
\newcite{sigir2019bert} presents a method for using 
question similarity and BERT-based 
question-answer relevance in FAQ retrieval. 
\newcite{xie2019faqbased} uses a knowledge graph-based Q\&A framework to improve questions understanding and answers retrieval, and evaluates it on a new community FAQ dataset\footnote{Not available yet.}.
Some recent works explore customer use of FAQs in the context of e-commerce\footnote{Not publicly available.}: \newcite{cui-etal-2017-superagent} shows how to search for customer questions in existing FAQ pairs.~\newcite{Kulkarni:2019} focuses on deep learning-based ranking model and ontology-based matching to answer question on products. All these previous works use community FAQ datasets (either public or not), whereas our work focuses on organizational FAQs, and differences between them and user questions.

Most relevant to our work is~\newcite{faq-nfaq} which studies community FAQs and user questions. They collected around 3000 questions from forums into FAQ and non-FAQ classes. Then, they analyzed different features, and found that named entities and personal pronouns are good discriminators between the classes.
The main difference with our work is that they focused on community FAQs, while we focus on organizational FAQs, and we further study the differences between community and organizational FAQs.

The task of FAQs generation was studied in previous works~\cite{Auto-FAQ-Gen,FAQtor}. Those works aim at generating community FAQs from online forum whereas we target organizational FAQs.

Evaluation of different aspects of pragmatics are explored in several other works. ~\newcite{rao-tetreault-2018-dear} studies differences between formal and informal questions.
Aspects such as informativeness and implicatures are presented in~\newcite{SQUINKY,Lahiri:2011}.

Recently, it is worth noting some relevant questions datasets related to the COVID-19 pandemic. CovidQA~\cite{tang2020rapidly} comprises 124 questions associated with answers from scientific articles, and the task is to identify the answer passage within the article. The questions were created by medical experts (epidemiologists, medical doctors, and medical students), and were curated from scientific articles.
In COVID-Q~\cite{wei2020people} the authors present a set of 1690 questions about COVID-19 from different sources and clustered into various categories. Most of their questions are extracted from FAQ pages. Some questions were extracted from Quora, as well as by keyword search on search engines. Although some of their question sources are similar to ours, the focus of their work is different.

\section{FAQ Analysis}
\label{faqs}

\subsection{User question vs. Organizational FAQ}
\label{user-corporate}
To study the differences between organizational FAQs and user questions, and since there is no publicly available dataset of organizational FAQs, we collected a large dataset of the two classes. To do so, we made use of the FAQPage schema\footnote{\url{https://schema.org/FAQPage}} defined by the Schema.org community,
that publishes standard schemas for structured data on the web.
This schema is used by organization experts to specify FAQs and answers in their 
websites, and thus, 
can be treated as ground-truth organizational FAQs.
We collected $19474$ 
FAQs from $2514$ organization websites. 
For the user question class, we used existing datasets that include user questions from either forums or search logs. Those questions were all written by users, not representing any organization, and thus represent ground-truth for user questions. In order to create a balanced dataset, we randomly selected around $3200$ questions from Quora\footnote{\url{https://tinyurl.com/yx2o2uo3}}, Yahoo! Answers L5\footnote{\url{https://tinyurl.com/thtjgpf}}, Reddit\footnote{\url{https://tinyurl.com/sl2uq38}}, forums including Ubuntu, Stackoverflow, Stats and Unix~\cite{ecir2019-1}, and WikiQA~\cite{wikiqa-emnlp} datasets. The data collection process ensures that both classes contain questions from various domains.

We train a \textsc{Question-type} classifier as follows:
we split the collected data into training (60\%), validation (20\%) and test (20\%) sets, and train an RNN model as in~\cite{howard-ruder-2018-universal}, using a pre-trained, 300-dimensional GloVe word embedding~\cite{pennington2014glove}. The hyper-parameters of the training are: RNN hidden layer size is 256, dropout is not applied, Adam optimizer is used with learning rate of 3e-5 and batch size is 256.
This classifier's accuracy on the test set is $91.8\%$, implying that there is indeed an inherent difference between the two classes: organizational FAQs and user questions.

\subsection{Community FAQ vs. Organizational FAQ}
\label{community_vs_organizational}

Given that we are able to detect organizational FAQs, we next try to evaluate whether community FAQs indeed differ from organizational FAQs.
Existing FAQ datasets contain FAQs, to which related user questions are matched. 
\textsc{FAQIR} is composed of $4313$ FAQs and $1233$ additional user questions in the ``maintenance \& repair'' domain, taken from Yahoo! Answers. \textsc{StackFAQ} is composed of $719$ FAQs and $1250$ user questions in the ``web app'' domain, taken from the StackExchange website. 
In both datasets, the sources of all FAQs are web forums. Hence, these are community FAQs.
We evaluate these community FAQs using the following metrics: \newline
\textbf{Organizational FAQ Percent.}
Using our \textsc{Question-type} classifier
, we predicted how many of the FAQs in the two datasets were likely to be organizational FAQs, and report the percentage of them. \newline
\textbf{Formal Percent.}
We trained an RNN classifier (with same hyper-parameters as described for the \textsc{Question-type} classifier) on the \textsc{Squinky} dataset~\cite{lahiri2015squinky}, which includes sentences labeled by their formality level, and used it to classify the FAQs in each dataset\footnote{Similarly to~\cite{pavlick2016empirical}, we consider sentences with formality level $> 3.75$ as ``formal'', and those with formality level $< 3.25$ as ``informal''.}. We report the percentage of FAQs classified as ``formal''. \newline
\textbf{Grammar Errors.}
Following~\newcite{kantor2019learning}, we used the Grammarly\footnote{\url{app.grammarly.com}} grammatical error correction system in order to count the number of errors per FAQ. We report the average number of grammatical errors. \newline
\textbf{Readability.} 
We used textstat\footnote{\url{https://pypi.org/project/textstat/}} implementation for Flesch-Kincaid grade level (F-K grade)~\cite{kincaid1975derivation} to assess readability of the FAQs, and report mean grade. The common wisdom for web content is average F-K grade of up to 7\footnote{\url{https://tinyurl.com/we6om8c}}, and on social media it is reported to be lower (4 on BuzzFeed, and 2.5 on Facebook)\footnote{\url{https://tinyurl.com/qnl5naa}}. \newline

\begin{table}[thpb]
\centering
\resizebox{\columnwidth}{!}{%
\begin{tabular}{l|c|c}
Metric & \textsc{FAQIR} & \textsc{StackFAQ} \\ \hline \hline
Organizational FAQ Percent & 7 & 39 \\
Formal Percent & 44 & 86 \\
Grammar Errors (mean) & 1.70 & 0.17 \\
Readability F-K Grade (mean) & 6 & 4.86 \\
\hline
 \hline
\end{tabular}
}
\caption{Descriptive metrics for community FAQs.}
\label{tab:CommunityFAQresults}
\end{table}

Table~\ref{tab:CommunityFAQresults} summarizes the evaluation of \textsc{FAQIR} and \textsc{StackFAQ} datasets. In both, a low percentage of community FAQs were classified as organizational FAQs. 
In \textsc{FAQIR} the majority of questions are not formal, and grammar error rate is high. Both datasets, especially \textsc{StackFAQ}, suffer from poor FAQ readability.
This analysis emphasizes the need of a new dataset that better models a professional user-support setting: users pose questions in their own words, and such questions should be matched against organizational FAQs in order to find appropriate answers and, on the other hand, frequent user-questions can be gathered and used to generate organizational FAQs.

\section{The \textsc{orgFAQ} Dataset}
\label{data}

\subsection{Data Collection}
\textsc{orgFAQ} is composed of organizational FAQs and user questions.
Collecting FAQs from organization websites that have enough corresponding user questions in community sources is a major challenge.
Since organizational FAQs are usually focused on a specific services or products, it is very difficult to find corresponding user questions, as communities typically focus only on a small set of very popular products. Interestingly, many organizations maintain an FAQ page that focuses on Jobs.
Such FAQ pages include questions which are general and common across organizations, e.g.: ``What should I expect in my interview?''.
This makes the Jobs domain an appropriate source from which a sufficiently large dataset can be drawn.
Thus, we collected job-related FAQs from $170$ organizations' websites, yielding a total of $1688$ FAQ questions and answers.
In parallel, 
we found two relevant sub-reddits ``/r/jobs/'' and ``/r/careerguidance/'' where users discuss 
related recruitment issues.
In total, around $134K$ user questions were collected.
All organization names appearing in the obtained data were replaced by a special ``ORG\_NAME'' token.
 
\subsection{Dataset Creation}
Given the sets of collected organizational FAQs and user questions, our goal was to create a high-quality dataset in which each organizational FAQ is associated with one or more corresponding user questions. Note that not each organizational FAQ in the collected data had necessarily a matching user question, and vice versa. Thus, we created the \textsc{orgFAQ} dataset in two main stages, namely automatic annotation followed by human annotation, as described in the following sub-sections.

\subsubsection{Automatic Annotation}
We created automatically-labeled data by associating each of the collected FAQs with three user questions. This was done as follows: \newline
\textbf{Step 1.} The entire set of $134K$ user questions was indexed in a full-text index. For each organizational FAQ, up to 10 best-matching user questions were retrieved from this index. The FAQ question itself served as the query, and the BM25 \cite{robertson2009probabilistic} metric was used for the retrieval. \newline
\textbf{Step 2.} We wrote a simple software tool for manually labeling the data. The tool presents to the labeler a series of organizational FAQs, each with the (up to 10) best-matching candidate user questions, according to the BM25 metric. Now, the human labeler annotates each candidate user questions as matching or not. This way, the authors of this paper manually labeled 1521 pairs, out of which, 268 pairs were labeled as matching. \newline
\textbf{Step 3.} Using the labeled data obtained in the previous step, we fine-tuned a pre-trained BERT model \cite{devlin2018bert}, for the task of classifying whether a pair of an organizational FAQ and a user question conveys the same meaning or not. \newline
\textbf{Step 4.} Finally, the fine-tuned BERT model was used for drawing the top-ranked user questions, per organizational FAQ. Due to running time considerations, we first used BM25 to retrieve the top-ranked 1000 user questions, for each organizational FAQ. We then re-ranked the retrieved user questions using the fine-tuned BERT model, and selected the top-three questions.

\subsubsection{Human Annotation}

The automatically-labeled data created at the previous step might be noisy. To ensure the quality of the dataset, we defined the following crowdsourcing task: each organizational FAQ was presented to a crowd annotator along with the three retrieved candidate user questions associated with it at the previous step.
For each candidate pair of FAQ and user question, the annotators had to decide whether they convey the same meaning, and if not, they had to rewrite the user question with minimal changes, so that it will match the FAQ. Asking the annotators to rephrase the user question, rather than writing a new one, carries a few benefits: writing requires more creativity than rephrasing, and hence, might result in sentences of low diversity, or very simple ones~\cite{jiang-etal-2017-understanding}. In addition, we wanted to retain the language style of the original user question (including syntax, typos, grammar, etc.) as much as possible. 
Each pair of FAQ and user question was annotated by three crowd annotators. We used the Appen platform\footnote{\url{https://appen.com/}}, and in order to ensure high quality data, we required only English native speakers who belong to the Highest Quality group (smallest group of most experienced, highest accuracy contributors). The inter-rater reliability, measured by Fleiss's Kappa, 
was $0.76$, indicating a high level of agreement. In addition, we randomly sampled $10\%$ of the pairs to be evaluated by two of the authors of this paper. The authors have found $88\%$ of the sentences to be valid, verifying the high quality of the data. In total, $6988$ pairs were created, and the number of user questions mapped to the same FAQ varies between $1$ and $10$, with an average of $4.4$.

\subsection{Dataset Analysis}
Analysis of the organizational FAQs and user questions in the \textsc{orgFAQ} dataset are summarized in Table~\ref{tab:statFAQinc} and Table~\ref{tab:CorFAQresults}.

Table~\ref{tab:statFAQinc} provides statistics about the dataset w.r.t vocabulary, length, etc. The first row captures the question type distribution, by showing the first word distribution for words with frequency above 2\% (for example, 20.5\% of the FAQs start with 'how'), which sums up to 91.7\%, and 71.2\% for the organizational FAQs, and user questions, respectively. This can indicate a more coherent and focused language style used in the organizational FAQs. In addition, while the mean question length is higher for the organizational FAQs, the vocabulary size is smaller, which also supports the indication above.
\begin{table}[t]
\centering
\resizebox{\columnwidth}{!}{
\begin{tabular}{l|l|l}
 & Organizational FAQs & User questions \\ \hline \hline
 \multirow{7}{*}{First word distribution} & 'how':20.5   & 'how':21.5 \\
 &'what':18.9 &'what':15.6 \\
 &'i':12.0 &'should':8.1\\
 &'can':9.0 &'is': 6.8\\
 {(above 2\%, in\%)}&'do':7.5 & 'can': 5.7\\
 &'is':3.9& 'i': 5.1 \\
 &'will':3.6 &'do': 3.2 \\
 &'when':3.0 &'when': 2.9 \\
 & 'does':3.0 & 'where': 2.3\\
 & 'where': 2.7 & \\
 &'if': 2.5& \\
 &'why': 2.1& \\
 &'are': 2.1 &\\\hdashline
 Total (above 2\%, in\%) & 91.7& 71.2\\\hline 
Vocabulary size (unique words) & 1805 & 2538 \\\hline
Mean question length (words) & 11.99 & 10.08 \\
\hline
 \hline
\end{tabular}
}
\caption{Statistics of \textsc{orgFAQ} dataset.}
\label{tab:statFAQinc}
\end{table}

Table~\ref{tab:CorFAQresults} shows evaluation of the dataset using the metrics described in section~\ref{community_vs_organizational}.
\begin{table}[t]
\centering
\resizebox{\columnwidth}{!}{%
\begin{tabular}{l|c|c}
Metric & Organizational FAQs & User questions \\ \hline \hline
Organizational FAQ Percent & \textbf{79} & 60  \\
Formal Percent & \textbf{90} & 74 \\
Grammar Errors (mean) & \textbf{0.08} & 0.30 \\
Readability F-K Grade (mean) &\textbf{8.2} & 6.62 \\
\hline
 \hline
\end{tabular}
}
\caption{Descriptive metrics for \textsc{orgFAQ} dataset.}
\label{tab:CorFAQresults}
\end{table} 
Collecting FAQs from organizations' FAQ pages resulted in a high percentage ($79\%$) of FAQs classified as organizational FAQs\footnote{To avoid
exposure to the Jobs domain, FAQ pages whose URLs included words such as jobs, career, etc. were excluded from the training set of the \textsc{Question-type} classifier.}.
The organizations' attention in creating their FAQs is reflected in a high formality percentage (90\%) and a very low grammar error rate (0.08). In addition, their average readability level ($8.2$) is significantly higher than the ones of \textsc{FAQIR} and \textsc{StackFAQ}. Such readability level corresponds to experts' recommendation for business content to be higher than 8\footnote{\url{https://tinyurl.com/y9mtysnb}}.

Finally, the user questions in our dataset are shown to be of lower quality than the FAQs in all metrics, as expected.

 \subsection{Dataset Samples}

Table~\ref{fig:examples} shows some representative examples from our dataset.
Note that several user questions could be mapped into the same FAQ question. For example, in the first row, there are four different user questions mapped to a single FAQ. We provide the FAQ answers (although it is out of this papers' scope, and is left for future work) for the completeness of the dataset.

\begin{table}[t]
\resizebox{1.0\columnwidth}{!}{
\scriptsize
\begin{tabular}{ |p{4cm}|p{2cm}|p{2cm}| }
\hline
\textbf{User questions} & 
\textbf{FAQ question} &
\textbf{FAQ answer} 
\\\hline

\begin{itemize}[label={-},after=\strut]
\item how to apply for position on your website that is not currently available 
\item am still really being considered for the position posted on your website
\item still waiting on an interview for a position posted on your website 
\item how to know if a position is still available?
\end{itemize} &
How do I know if a position posted on your website is still available? &
We post all open positions on our website. We remove those positions once they are filled, canceled or put on hold.
\\\hline
\begin{itemize}[label={-},after=\strut]
\item what is the minimum age to work in the company?
\item what minimum age should i have to apply?
\item minimum age to work
\item what is the minimum age to get the job? 
\end{itemize}&
What’s the minimum age requirement to work at ORG\_NAME? &
The minimum age to be eligible for employment at ORG\_NAME is 18 years old. However, many positions require a 21-year-old minimum.
\\ \hline
\begin{itemize}[label={-},after=\strut]
\item should apply again after being rejected
\item i can run again, i was previously rejected
\end{itemize} &
I have applied before and was rejected. Should I try again? &

ORG\_NAME recruits on a post-by-post basis, so your application will only have been for the particular post advertised. This means you are free to apply again.
\\\hline
\end{tabular}}
  \caption{Examples from the \textsc{orgFAQ} dataset.}~\label{fig:examples}

\end{table}

\section{Experiments}

In this section, we show how the \textsc{orgFAQ} dataset can be used through the task of ``Organizational FAQ Generation from User Questions'': given a set of user questions with similar meaning, the goal is to produce a corresponding organizational FAQ. With such a capability, an organization can gather user questions on the web, assign them to clusters (e.g. by semantic similarity), and generate an appropriate FAQ for each cluster. We first show how \textsc{orgFAQ} can be used for this task in the domain of jobs, and then how this can be generalized to another domain.

\subsection{Organizational FAQ Generation for the Jobs Domain}
\label{experiment}

Naturally, we can use the \textsc{orgFAQ} dataset for training an FAQ generator in the jobs domain. To do so, we used the abstractor neural network of~\newcite{fast2018chen}, a sequence-to-sequence model with attention and copy mechanism. A training sample consists of a set of up to 10 user questions (4.4 on average) as the input, and the organizational FAQ as the target. The user questions were concatenated into a single sequence of tokens, with a special token separator.
We split a total of 1579 samples into training (85\%), validation (5\%) and test (10\%) sets. Default training hyper-parameters were used, and the validation set was used for early stopping. 
Using ROUGE metrics~\cite{lin2004rouge}, we compare our model's performance against a baseline method which randomly selects one of the input user questions. The results, shown in Table~\ref{tab:RougeResults}, are mean results over ten experimental rounds. On each round, a generation model was trained and evaluated using a different random training/validation/test split. Table~\ref{tab:GenerationExample} shows examples of a generated FAQs, given input user questions. The first example, (A), shows a high-quality output - it demonstrates the ability of the model to convert several user questions with different kinds of errors (grammatical errors, missing question mark, ``when'' instead of ``where'') into a properly-phrased question which can form an adequate organizational FAQ. The second example, (B), shows a lower-quality output FAQ, where the subject of the sentence is wrong.

\begin{table}[htbp]
\centering
\resizebox{\columnwidth}{!}{%
\begin{tabular}{l|c|c|c}
Method & ROUGE-1 & ROUGE-2 & ROUGE-L \\ \hline \hline
Baseline & 0.46 & 0.25 & 0.43 \\ \hline
FAQ-Generator & \textbf{0.53} & \textbf{0.33} & \textbf{0.50} \\ \hline
\end{tabular}
}
\caption{Mean ROUGE $F_1$ scores of our FAQ-Generator vs. a baseline method which randomly selects one of the input user questions.}
\label{tab:RougeResults}
\end{table}

\begin{table}[htbp]
\centering
\resizebox{\columnwidth}{!}{%
\begin{tabular}{|l|l|}
\hline
\multirow{2}{*}{User questions}
& when are you located ? \\
& where is the job \\
& how do answer where are you located \\
& where should located \\
& where they are \\
\hline
Generated FAQ & where are you located ? \\ \hline
Target FAQ & where are you located ? \\ \hline
\multicolumn{2}{c}{(A) High-Quality Generation } \\ 
\multicolumn{2}{c}{ } \\
\hline
\multirow{2}{*}{User questions}
& can i ask how the salary is paid ? \\
& is it okay to ask how the salary is paid ? \\
& how do you pay your salaries ? \\
& how do i talk about mode of pyment of salary ? \\
\hline
Generated FAQ & how do i pay your salaries ? \\ \hline
Target FAQ & how is salary paid ? \\ \hline
\multicolumn{2}{c}{(B)  Low-Quality Generation } \\
\end{tabular}
}
\caption{Examples of FAQs generated by the model, given input user questions. Also shown are the targets (ground-truth FAQs).}
\label{tab:GenerationExample}
\end{table}

\subsection{Generation for a new Domain: the COVID-19 Use Case}
\label{covid}

Here, we present a use-case that utilizes the \textsc{orgFAQ} dataset for a new domain, showing its usefulness.
\subsubsection{Setting}
The COVID-19 pandemic, also known as the coronavirus pandemic was first identified in December 2019, and quickly affected millions of people all over the globe: hence, information has become a critical aspect of the pandemic. People were frantically seeking information, ranging from ``what is the covid-19 virus?'' to ``can pets get the coronavirus?'', and more. People posted questions in different public social media channels such as Twitter, Quora, Reddit, etc. In addition, new dedicated channels (using emails, forms, etc.) were created by different organizations such as countries, municipalities, universities and companies. People were encouraged to use those channels for asking questions. For example, the city of Hillsboro, Oregon, USA provided a form to ask questions about the Covid-19\footnote{\url{https://tinyurl.com/y9ftbpbs}}.
Practically, those organizations used questions from different channels in order to create and update their COVID-19 FAQs pages.
An FAQ-Generator, as described in section~\ref{experiment}, can help automating this process and reduce human effort. Given the lack of a sufficiently large training data, we explored the benefit of utilizing a pre-trained model trained on a different domain, such as FAQ-Generator trained on the \textsc{orgFAQ} dataset, as a starting point for further training on a small COVID-19 dataset.

\subsubsection{Questions Collection}

As before, the inputs to the FAQ-Generator should be user questions clustered by similar topics.
In order to create such clusters, and given the limited number of publicly available user questions, our approach was to define topics, and to use COVID-19 dedicated social media sources, namely Quora ``Shared knowledge and experiences'' about COVID-19\footnote{\url{https://tinyurl.com/ybnm4b5k}} and, Kaggle Coronavirus (covid19) Tweets dataset\footnote{\url{https://tinyurl.com/yb5znxcr}}, to find user questions which are relevant for each topic. In addition, we asked colleagues to write additional relevant user questions.
In total, we created a set of 57 topics, and an average of 13.7 user questions per topic. Finally, we created ground-truth labels for each topic by searching FAQs in official websites like the CDC (Centers for Disease Control and Prevention)\footnote{\url{https://tinyurl.com/vwxtxpp}}, WHO (World Health Organization)\footnote{\url{https://tinyurl.com/y8674r39}} and more. An average of 4.3 FAQs were collected per topic. 
Table~\ref{tab:covid-topic-example} shows an example of collected questions on the topic ``Blood donation''.

\begin{table}[htbp]
\centering
\resizebox{\columnwidth}{!}{%
\begin{tabular}{|l|l|}
\hline
\multirow{2}{*}{User questions}
& I'm not sick with coronavirus can donate? \\
& is it okay to donate blood \\
& are the any evidences of virus transmission thru blood \\
& can COVID19 be transmitted thru blood\\ \hline
FAQ questions & Can covid-19 be transmitted by blood donation? \\
& Can covid-19 be transmitted by blood transfusion? \\\hline
\end{tabular}
}
\caption{An example of ``Blood donation'' topic, with four user questions and two ground-truth FAQs.}
\label{tab:covid-topic-example}
\end{table}

\subsubsection{COVID-19 FAQ Generation}
Out of our collected COVID-19 data, we prepare data for FAQ-Generator training, similarly as described in section~\ref{experiment}. Again, each training sample consists of a set of up to 10 user questions as input, and an organizational FAQ as target. A topic with $n$ user questions and $k$ FAQ questions contributes $k$ training samples, each with $min(n,10)$ user questions (in case $n>10$, we randomly select 10 user questions for each training sample). For ROUGE evaluation, all $k$ FAQ questions of a topic from the test set, serve as ground-truth references of the corresponding test sample. Thus, a rather small dataset of 245 samples is obtained out of the 57 topics. Splitting to training (80\%), validation (10\%) and test (10\%) sets was done at the topic level, to avoid same user questions appearing in both training and test (or validation) sets. Same neural network architecture and hyper-parameters were used as described in section~\ref{experiment}. The validation set was used for early stopping. As before, the reported ROUGE results are mean results over ten experimental rounds.

Here we would like to explore the benefit of leveraging a pre-trained model trained on the Jobs domain, given the insufficiently large COVID-19 dataset which we were able to collect. Hence, we compare the performance of two models: one trained solely on the COVID-19 data, while the other is initialized with the parameters of a model pre-trained on the \textsc{orgFAQ} dataset, and fine-tuned on the COVID-19 data. We additionally evaluate a model trained solely on the \textsc{orgFAQ} data, without fine-tuning, and a baseline method which randomly selects one of the input user questions. The results, shown in Table~\ref{tab:CovidRougeResults}, demonstrate the usefulness of our \textsc{orgFAQ} dataset. While the COVID-19 dataset by itself is too small to obtain a better model than the baseline, pre-training on the Jobs data significantly improves performance.
It implies that some of the knowledge learnt by the Jobs pre-trained model is generalizable to other domains.
However, pure transfer learning was not sufficient, and the fine-tuning stage was needed for obtaining improved results. Table~\ref{tab:CovidGenerationExample} shows examples of generated FAQs by the fine-tuned model. The first one, (A), exemplifies a concise and accurate output FAQ, in spite of some noise and irrelevant details in some of the input user questions. In the second example, (B), the generated FAQ is of lower quality, due to wrong polarity of the verb.

\begin{table}[htbp]
\centering
\resizebox{\columnwidth}{!}{%
\begin{tabular}{l|c|c|c}
Method & ROUGE-1 & ROUGE-2 & ROUGE-L \\ \hline \hline
Baseline & 0.33 & 0.14 & 0.31 \\ \hline
FAQ-Generator - \textsc{orgFAQ} only & 0.29 & 0.11 & 0.27 \\ \hline
FAQ-Generator - COVID-19 only & 0.27 & 0.08 & 0.26 \\ \hline
FAQ-Generator - COVID-19 fine-tuned & \textbf{0.37} & \textbf{0.17} & \textbf{0.36} \\ \hline
\end{tabular}
}
\caption{Mean ROUGE $F_1$ scores of FAQ-Generator pre-trained on the \textsc{orgFAQ} data and fine-tuned on the COVID-19 data, versus alternative models which lack pre-training or fine-tuninig. The baseline method randomly selects one of the input user questions.}
\label{tab:CovidRougeResults}
\end{table} 

\begin{table}[htbp]
\centering
\resizebox{\columnwidth}{!}{%
\begin{tabular}{|l|l|}
\hline
\multirow{2}{*}{User questions}
& where can i get tested in alabama \\
& i live in austin , where can i get tested for coronavirus \\
& where can i get tested in new mexico \\
& i live in ontario , where can i get tested ? \\
& how do i get a coronavirus test ? \\
& where can i go to get tested for coronavirus ? \\
& where i can get tested \\
& how do i get screened ? \\
& i need to be checked for coronavirus . \\
& how can i get a coronavirus test \\
\hline
Generated FAQ & how can i get tested ? \\
\hline
Target FAQ & where can i go to get tested ? \\
\hline
\multicolumn{2}{c}{(A) High-Quality Generation } \\ 
\multicolumn{2}{c}{ } \\
\hline
\multirow{2}{*}{User questions}
& why are there no masks and respirators in pharmacies ? \\
& how is it possible that the state does not have enough masks and respirators ? \\
& why can't i get or find a mask or a respirator anywhere ? \\
& when will masks be available ? \\
& why aren't there respirators ? \\
& why are no masks or respirators available anywhere ? \\
& i'd like a respirator \\
& where can i get masks ? \\
& where do people get protective masks or disinfectants ? \\
& when can i buy a mask at a pharmacy ? \\
\hline
Generated FAQ & why can i buy a mask ? \\
\hline
Target FAQ & is there a shortage of masks ? \\ \hline
\multicolumn{2}{c}{(B)  Low-Quality Generation } \\
\end{tabular}
}
\caption{Examples of FAQs generated by the COVID-19 fine-tuned model, given sets of input user questions. Also shown are the targets (ground-truth FAQs).}
\label{tab:CovidGenerationExample}
\end{table} 

\section{Conclusion}
\label{conclusions}

We study differences between community FAQs and organizational FAQs, propose a new dataset to unleash research in the field, and show its usefulness on a new task for different domains. For future work, we plan to use this dataset on more tasks, such as question answering, and look into the properties of FAQ answers.

\bibliography{references}
\bibliographystyle{acl_natbib}


\end{document}